\documentclass{article}

\usepackage[preprint, nonatbib]{neurips_2018}

\usepackage[utf8]{inputenc} 
\usepackage[T1]{fontenc}    
\usepackage{hyperref}       
\usepackage{url}            
\usepackage{booktabs}       
\usepackage{amsfonts}       
\usepackage{nicefrac}       
\usepackage{microtype}      

\usepackage{amsmath}
\usepackage{algorithm, algorithmic}
\usepackage{amsfonts, amssymb}
\usepackage{multirow}
\usepackage{graphicx} 

\title{Automatic Induction of Neural Network Decision Tree Algorithms}

\author{%
  Chapman Siu \\
  Faculty of Engineering and Information Technology\\
  University of Technology Sydney\\
  Sydney, Australia \\
  \texttt{chapman.siu@student.uts.edu.au} \\
}

\begin{document}

\maketitle

\begin{abstract}
This work presents an approach to automatically induction for non-greedy decision trees constructed from neural network architecture. This construction can be used to transfer weights when growing or pruning a decision tree, allowing non-greedy decision tree algorithms to automatically learn and adapt to the ideal architecture. In this work, we examine the underpinning ideas within ensemble modelling and Bayesian model averaging which allow our neural network to asymptotically approach the ideal architecture through weights transfer. Experimental results demonstrate that this approach improves models over fixed set of hyperparameters for decision tree models and decision forest models.
\end{abstract}
\section{Introduction}

Decision trees and their variants have had a rich and successful history in machine learning in general, and their performance has been empirically demonstrated in many competitions and even in automatic machine learning settings.

Various approaches have been used to enable decision tree representations within a neural network setting, in which this paper will consider non-greedy tree algorithms which are built on top of oblique decision boundaries through probabilistic routing\cite{nongreedy}. In this way, decision tree boundaries and the resulting classification is treated as a problem which can be learned through back propagation in a neural network setting \cite{dndf}.

On the neural network component, it has been further demonstrated that highway networks can viewed as an ensemble of shallow neural networks \cite{veit}. As ensembles of classifiers are related to the Bayesian Model Averaging in an asymptotic manner \cite{le2017}, thus, creating a decision tree model within a neural network setting over a highway network can be used to determine the optimal neural network architecture and by extension the optimal hyperparameters for decision tree learning. 

As our contribution, we aim to provide an automated way to induce decision tree whilst retaining existing weights in order to progressively grow or prune decision trees in an online manner. This simplifies the hyperparameters required in choosing models, instead allowing our algorithm to automatically search through the ideal neural network architecture. As such in this work, we modify existing non-greedy decision tree algorithms through stacking our models through modifying the routing algorithm of decision trees. Thus whilst previously, a single induced decision tree may have only one set of training for the leaf nodes, in our approach, a single decision tree can have different set of leaf nodes stacked in order determine the ideal neural network architecture. 

This paper seeks to identify and bridge two commonly used Machine Learning techniques in the form of tree models and neural networks, as well as identifying  some  avenues  for  future  research .


\section{Background}

Within decision tree algorithms, research has been done to grow tree and prune trees in a post-hoc manner, greedy trees are limited in their ability to fine tune the split function once a parent node has already been split\cite{nongreedy}. In this section, we will briefly outline related works for non-greedy decision trees, approaches to extending non-greedy tree to ensemble models and finally mechanisms for performing model choice through Bayesian model determination.

\subsection{Inducing Decision Trees in Neural Networks}


Inducing non-greedy decision trees has been done through construction of oblique decision boundaries \cite{nongreedy}\cite{dndf}. This has been done through soft-routing of the decision tree wherein the contribution of each leaf node to the final probability is determined probabilistically. One of the contributions by Kontschieder et al\cite{dndf} compared with other approaches the separating of training the probabilistic routing of the underlying binary decision tree and the training of the leaf classification nodes which need not be binary classification. The decision tree algorithm was also modified in through a shallow ensemble manner to a decision forest through bagging the classifiers. 

In early implementation of decision trees, algorithms used were often using recursive partitioning methods, which aim to perform partitions in the form of $X_i > k$ where $X_i$ is one of the variables in the dataset and $k$ is a constant which is the split decision. These decision trees are also called \textit{axis-parallel}\cite{Murthy1994oblique}, because each node produces a axis-parallel hyperplane in the attribute space. These trees are often considered greedy trees, as they grow a tree one node at and time with no ability to fine tune the splits based on the results of training at lower levels of the tree\cite{nongreedy}.

In contrast, recent implementations of decision trees focus instead on the ability to update the tree in an \textit{online} fashion leading to non-greedy optimizations typically based on \textit{oblique decision trees}\cite{nongreedy}\cite{dndf}. the goal of oblique decision trees is to change the partition decisions instead to be in the form $\sum_{i=1}^p a_i X_i + a_{p+1} > 0$ where $a_i, ..., a_{p+1}$ are real-valued coefficients. Theses tests are equivalent to hyperplanes at an oblique orientation relative to the axis hence the name \textit{oblique decision trees}. From this setting, one could convert oblique decision trees to the axis-parallel counterpart by simply setting $a_i = 0$ for all coefficients except one. 

\subsection{Ensemble Modelling and the Model Selection Problem}

Ensemble modelling within the neural networks has also been covered by Veit et al. \cite{veit}, who demonstrated the relationship between residual networks (and by extension Highway Networks) and the shallow ensembling of neural networks, in the form $y_i^{(1)} \cdot t(y_i^{(1)}) + y_i^{(2)} \cdot t(y_i^{(2)}) + ... + y_i^{(n)} \cdot t(y_i^{(n)}) $. Furthermore, in this setting as we are interested in stacking models of the same class, Le and Clarke \cite{le2017} have demonstrated the asymptotic properties in stacking and Bayesian model averaging. 
Approaches like sequential Monte Carlo methods\cite{green} can be used in order in order to change state and continually update the underlying model. 

A simple approach to consider an ensemble approach to the problem. In this setting we would simply treat the new data \textit{independent} of the old data and construct a separate learner. Then we can combine it together using a stacking approach. In this setting, we aim to combine models as a linear combination together with a base model which might represent any kind of underlying learner \cite{Wolpert1992}. 

More recently there have been attempts at building ensemble tree methods for online decision trees, including the use of bagging techniques in a random forest fashion\cite{dndf}. Furthermore it has been demonstrated that boosting and ensemble models have connections with residual networks\cite{HuangALS17}\cite{veit}, giving the rise to the possibility of constructing boosted decision tree algorithms using neural network frameworks.

These approaches to ensemble models have a Bayesian parallel. In the Baye-sian model averaging algorithms. These models are related to stacking\cite{le2017}, where the marginal distribution over the data set is given by $p(\mathbf{x}) = \sum_{h=1}^H p(\mathbf{x} \vert h) p(h)$. The interpretation of this summation over $h$ is that just one model is responsible for generating the whole data set, and the probability distribution over $h$ reflects uncertainty as to which model that is. As the size of the data set increases, this uncertainty reduces and the posterior probabilities $p(h\vert \mathbf{x})$ become increasingly focused on just one of the models\cite{bishop2006}.

\section{Our Approach}

In this section we present the proposed method in which we describe our approach to automatically grow and prune decision trees. This section is divided into the following parts: decision routing, highway networks and stacking.

\subsection{Decision Routing}

\begin{figure*}[htp]
\centering
\includegraphics[width=\textwidth]{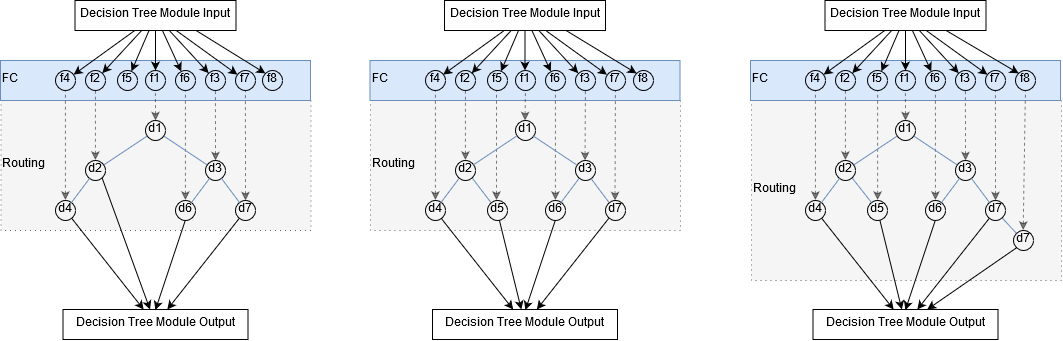}
\caption{Multiple decision routing for the same shared fully connected layers. Left to right: tree of depth two with one node pruned, tree of depth two with no perturbation, tree of depth two with one node grafted. The fully connected (FC) block (in blue) is shared among all the trees, through the parameters are not updated equally if they are not routed as part of the routing algorithm. When constructed in a neural network and stacked together, the network weights would only comprise of the rightmost structure (with the additional node pruned), with multiple outputs representing each of the routing strategies. At each of the leaf nodes, there would be a separate classifier layer that is built in accordance with the number of classes which the decision tree is required to train against.}
\label{fig:dr}
\end{figure*}

In our decision routing, a single neural network can have multiple routing paths for the same decision tree. If we start from the base decision tree, we could have two additional variations, a tree with one node pruned and a tree with one additional node grafted. In all three scenarios, the decision tree would share the same set of weights; the only alteration is that the routing would be different in each case as shown in Figure \ref{fig:dr}. 

In this scenario, all trees shared the same underlying tree structure and were connected in the same way. I it is in this manner which weights can be shared among all the trees. The routing layer determines whether nodes are to be pruned or grafted. The decision to prune or graft a node was done through $p(x_{t+1} | x_t, \theta)$. In the simpliest case, we simply pick a leaf node uniformly at random to prune or to graft. Additional weighting could be given depending on the past history of the node and updated using SMC approaches with a uniform prior. 

\subsection{Highway Networks and Stacked Ensembles}

In order to enforce stacking as a highway network, the function $t$ would be weights of one dimension, that is that that it is a scalar of one dimension, that is $t^j(y_i, \theta^j)$ is $\theta^j \in \mathbb{R}$ for all $j$ where $\sum_{\forall j} \theta ^j = 1, \theta^j \geq 0$, $\forall \theta^j$. 

\begin{figure}[htp]
\centering
\includegraphics[width=8cm]{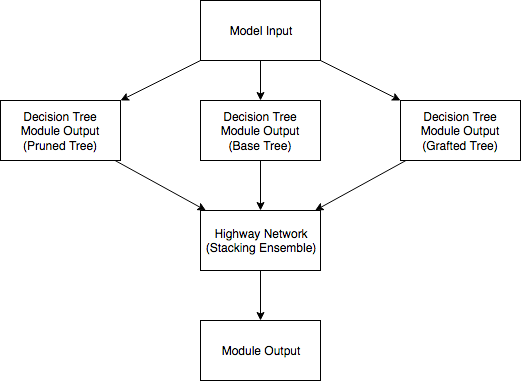}
\caption{Stacking Ensembles can be achieved through the use of a Highway Network}
\label{fig:dh}
\end{figure}

In this manner, the different decision trees which are perturbed as a stacked ensemble as shown in Figure \ref{fig:dh}. Using this information, the corresponding weights can be interpreted in the Bayesian model sense. 

The construction of such a network differs from the usual highway network in the sense that the underlying distribution of data does not alter how it is routed, instead all instances are routed in the same way which is precisely how the stacking ensembles operate, as oppose to the usage of other ensemble methods. The advantages of using a simple stacking in this instance, is primarily the interpretation of the weights as posterior probability of the Bayesian model selection problem. 

\subsection{Model Architecture Selection}

Finally the model architecture selection problem can be constructed through combining the two elements above. In this setting, at every iteration we would randomly select nodes to prune and grow. At the end of each iteration, we would perform weighted random sampling based on the model posterior probabilities. 

After several iterations we would expect that $p(h| x_t)$ will eventually converge to a particular depth and structure of the decision tree. In order to facilitate this, the slope annealing trick $\pi = \text{softmax}_\tau(y/\tau)$\cite{chung2017}, where $\pi$ is the modified weighted samples, and $y$ is the output from the highway network and $tau$ is the temperature. This is introduced to the highway net weights in order to progressively reduce the temperature so that base model selected to perturb becomes more deterministic in the end.

\begin{algorithm}
\caption{Automatic Induction of Neural Decision Trees}
\begin{algorithmic}[1]
\STATE randomly initialize $\theta$
\FOR{$t=1$ to $T$}
\STATE Based on $p(h | x_t, \theta_t)$ (i.e. the weights inferred by the Highway Network), draw a weighted sample to perturb the model $x_t$ based on $\text{softmax}_\tau$
\STATE Compute and update the weights $\theta$ using back propagation
\STATE Lower temperature of softmax to $\tau = \tau \times \delta$, where $\delta$ is the discount rate
\ENDFOR
\STATE Return final model $x_t$ with probabilistic routing and parameters $\theta_t$
\end{algorithmic}
\end{algorithm}

Furthermore, this can be extended to ensemble approach through construction of such trees in parallel leading to the decision forest algorithms. In this scenario, each tree in the forest will have its own set of parameters and will induce different trees randomly. As they would be induced separately and randomly, we may yield more diverse set of classifiers leading to stronger results which may optimize different portions of the underlying space.

\section{Experiments and Results}

In our experiments we use ten publicly available datasets to demonstrate the efficacy of our approach. We used training and test datasets where provided to compare performance as shown in the table below. Where not provided, we performed a random $70/30$ split into training and testing respectively. The following table reports the average and median, over all the datasets, the relative improvement in log-loss over the respective baseline non-greedy decision tree. 


\begin{table}
\caption{Average Error Improvement Across compared with baseline}
\begin{center}
\begin{tabular}{|l|c|c|c|c|}
\hline
Model & Avg Impr. & Avg Impr. & Median Impr. & Median Impr.\\
& (Train) & (Test) & (Train) & (Test) \\
\hline
Tree   & 8.688\%  & 6.276\% & 1.877\% & 0.366\% \\
Forest & 7.351\%  & 7.351\% & 0.247\% & 0.397\% \\
\hline
\end{tabular}
\end{center}
\end{table}

\begin{table}
\caption{Average Error Improvement Across compared with baseline, with fine tuning}
\begin{center}
\begin{tabular}{|l|c|c|c|c|}
\hline
Model & Avg Impr. & Avg Impr. & Median Impr. & Median Impr.\\
& (Train) & (Test) & (Train) & (Test) \\
\hline
Tree   & 22.987\% & 12.984\%  & 11.063\% & 4.461\% \\
Forest & 23.223\% & 15.615\%  & 11.982\% & 5.314\% \\
\hline
\end{tabular}
\end{center}
\end{table}

In all instances, we began training our decision tree with a depth of $5$, with $\tau$ initialized at $1.0$ with a discount rate of $0.99$ per iteration. Our decision forest was also set to have $5$ decision trees, and combined through average voting. For all datasets we used standard data preparation approach from the recipes R library whereby we center, scale, and remove near zero variance predictors from our datasets. All models for baseline and our algorithm were built and trained using Python 3.6 running Keras and Tensorflow. In all models we use decision tree of depth 5 as a starting point with benchmark models trained for 200 epochs. With out automatic induced decision trees we train our models for 10 iterations, each with 20 epochs. We further train the final selected models to fine tune the selected architecture with the results as shown.

From the results, we notice that both approaches improve over the baseline where the tree depth is fixed to $5$. With further fine tuning, it become apparent that the decision forest algorithm outperforms the vanilla decision tree approach. Even without fine tuning, it is clear that the forest approach is more robust in its performance against the testing dataset, demonstrating the efficacy of our approach.

\section{Conclusion}

From the results above, and compared with other benchmark algorithms, we have demonstrated an approach for non-greedy decision trees to learn ideal architecture through the use of sequential model optimization and Bayesian model selection. Through the ability to transfer learning weights effectively, and controlling the routing, we have demonstrated how we can concurrently train strong decision tree and decision forest algorithms, whilst inducing the ideal neural network architecture.

\bibliographystyle{plain}
\bibliography{main}


\end{document}